\documentclass[]{spie}  
\usepackage{graphicx}
\graphicspath{ {./} }
\usepackage{url}
\usepackage{amsmath}
\usepackage{amssymb}
\usepackage{xcolor}

\title{Conditional Generative Adversarial Networks for Data Augmentation and Adaptation in Remotely Sensed Imagery} 

\author{Jonathan Howe\supit{a}, Kyle Pula\supit{b}, Aaron A. Reite\supit{c}
\skiplinehalf
\supit{a}NVIDIA, 2788 San Tomas Expy., Santa Clara, CA, USA;\\
\supit{b}CACI, 15955 E Centretech Pkwy., Aurora, CO, USA;\\
\supit{c}NGA Research, 7500 GEOINT Dr., Springfield, VA, USA\\
}

\authorinfo{Further author information: (Send correspondence to J.H.)\\ J.H.: E-mail: jhowe@nvidia.com\\ K.P.: E-mail: jpula@caci.com\\ A.A.R.: E-mail: aaron.a.reite@nga.mil\\}

\begin{document} 
  \maketitle 

\begin{abstract}
The difficulty in obtaining labeled data relevant to a given task is among the most common and well-known practical obstacles to applying deep learning techniques to new or even slightly modified domains. The data volumes required by the current generation of supervised learning algorithms typically far exceed what a human needs to learn and complete a given task. We investigate ways to expand a given labeled corpus of remote sensed imagery into a larger corpus using Generative Adversarial Networks (GANs). We then measure how these additional synthetic data affect supervised machine learning performance on an object detection task.

Our data driven strategy is to train GANs to (1) generate synthetic segmentation masks and (2) generate plausible synthetic remote sensing imagery corresponding to these segmentation masks. Run sequentially, these GANs allow the generation of synthetic remote sensing imagery complete with segmentation labels. We apply this strategy to the data set from ISPRS' 2D Semantic Labeling Contest - Potsdam, with a follow on vehicle detection task. We find that in scenarios with limited training data, augmenting the available data with such synthetically generated data can improve detector performance. 

\end{abstract}


\keywords{Remote sensing, Deep Learning, Object Detection, Synthetic Data, Generative Adversarial Networks}

\section{Introduction}
\label{sec:intro}  

\begin{figure}[htbp]
\centering
\includegraphics{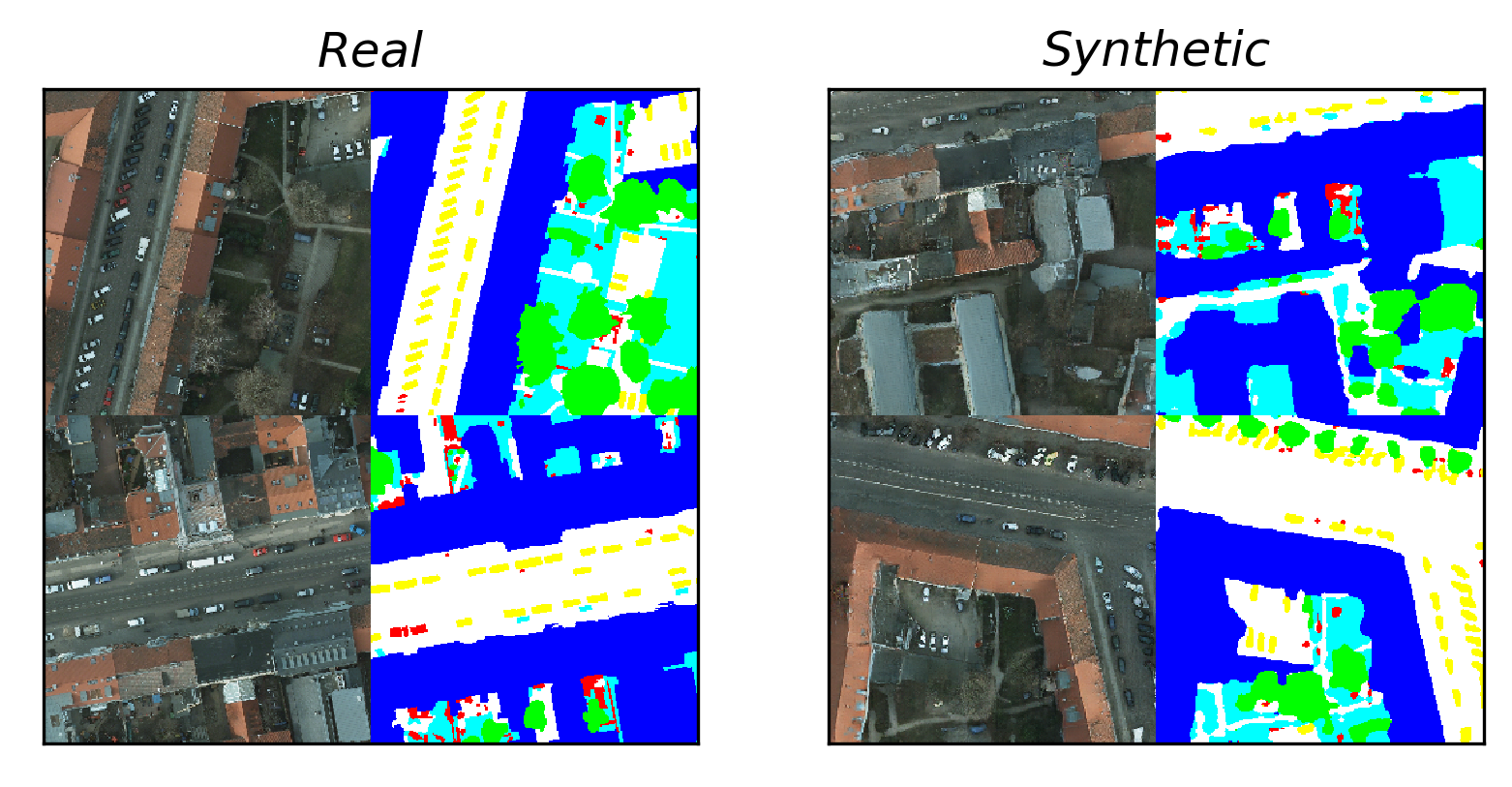}
\caption{Examples of real and synthetic image-label pairs when modeling the ISPRS Potsdam 2D Semantic Labeling Contest data set using a combination of Progressive and Conditional Generative Adversarial Networks \cite{isprs}. The labels include the categories of impervious surfaces (white), buildings (blue), low vegetation (aqua), tree (green), vehicle (yellow), and clutter/background (red).}
\label{intro_image}
\end{figure}

Developing a corpus of labeled data is imperative to train a deep learning capability in a supervised fashion to a sufficient accuracy for deployment. Unfortunately, access to such high quality labeled training data is often difficult, particularly when developing applications for new tasks, modalities, domains or classes. For remote sensing applications, there exist a number of open source data sets that can be used for model development \cite{xview,isprs,spacenet,dstlkaggle}. Although these data sets are extremely useful for research purposes, their size, scale and variety are usually insufficient to develop capabilities for real world applications. 

While there are a number of methods to create a labeled corpus for supervised training purposes, doing so efficiently, at scale and with extensibility in mind requires careful thought. Active learning and bootstrapping approaches in a collaborative environment, beginning from a small labeled data corpus, can help developers rapidly label data whilst also generate a deployable capability \cite{konyushkova2017learning}. Therefore, intelligent data augmentation approaches, particularly when the initial labeled data corpus is small, can be very beneficial in accelerating the development of a capability. 

There have been a number of successful attempts using data synthesis to augment the training data corpus. One avenue is to programmatically render objects and backgrounds, which allows the user to have full control over the scene and to obtain ground truth labels easily. Recent examples of this include the work by Tremblay et al where random textures and patterns are placed onto objects of interest, in this case vehicles for autonomous driving purposes \cite{tremblay2018training}. Rendered vehicles are then placed into random scenes, with the premise that salient features will be modeled for object detection purposes. This idea was extended in the work of Prakash et al where domain randomized objects and scenes were mixed with more contextually realistic scenes \cite{prakash2018structured}. Similar work has been performed for LIDAR data sets where 3D rendered vehicles are placed in the scene to pretrain an object detector  \cite{brekke2019multimodal}. For remote sensing applications, rendered maritime vessels have been placed into real imagery to vastly improve object detection metrics \cite{ijgi8060276}. One issue with physically rendering objects and backgrounds is the amount of time required to compose and create the scenes. In some cases it may not be possible, without significant investment, to perform this exercise.

The alternative approach considered in this paper is to use data driven techniques to model the underlying distribution and variation within the data set. This reduces the need to carefully construct and render objects and scenes. In addition, if the data is modeled sufficiently well, the composition and texture of the data can be extremely realistic. For example, Generative Adversarial Networks (GANs) can synthesize human faces to a degree where it is difficult for humans to differentiate GAN generated images from real images \cite{karras2017progressive, karras2018style}. Similar approaches have been used to augment small training data sets, particularly in the healthcare area \cite{lau2018scargan, sixt2018rendergan, shrivastava2017learning}. This approach has been found to be beneficial when the number of training data is low \cite{lesort2018evaluation}. Few attempts have been made to augment remote sensing data using these approaches, and those that have mostly focus on image classification \cite{lin2017marta, seo2018domain}. 

We develop a data driven strategy that trains a pair of progressive and conditional GANs to jointly model segmentation mask labels and corresponding remote sensing imagery from the publicly available data set provided by the International Society for Photogrammetry and Remote Sensing's (ISPRS) 2D Semantic Labeling Contest - Potsdam (see Figure \ref{intro_image}). We then study how augmenting available training data with such synthetically generated labeled data affects a vehicle detection task. The authors believe this is the first time such a joint image and label modeling and generation approach has been attempted in the context of detection. In scenarios where little real data is available (e.g., fewer than 400 vehicles), our augmentation strategy demonstrates a clear benefit. 

\section{Approach} 
\label{sec:approach}

\subsection{Data Set}

Our study used the ISPRS 2D Semantic Labeling Contest - Potsdam visible band (RGB) data set. The data set contains 24 segmented labeled images with six categories of object or land use type: impervious surfaces, buildings, low vegetation, tree, vehicle, and clutter / background. All images are collected at a nadir perspective from a low flying aircraft with a 6000 x 6000 image size and a 5cm ground sample distance (GSD)\cite{isprs}.

To emulate imagery obtained from a hypothetical low earth orbit electro-optical satellite with a 30cm GSD, the images and labels are down sampled using nearest neighbor interpolation producing 1000 x 1000 image sizes. The data set is split into training and test sets consisting of 20 and 4 image-label pairs respectively. The selected test images are; \textit{top{\_}potsdam{\_}4{\_}12}, \textit{top{\_}potsdam{\_}3{\_}12}, \textit{top{\_}potsdam{\_}5{\_}11}, and \textit{top{\_}potsdam{\_}7{\_}12}. 

\begin{figure}[htbp]
\centering
\includegraphics{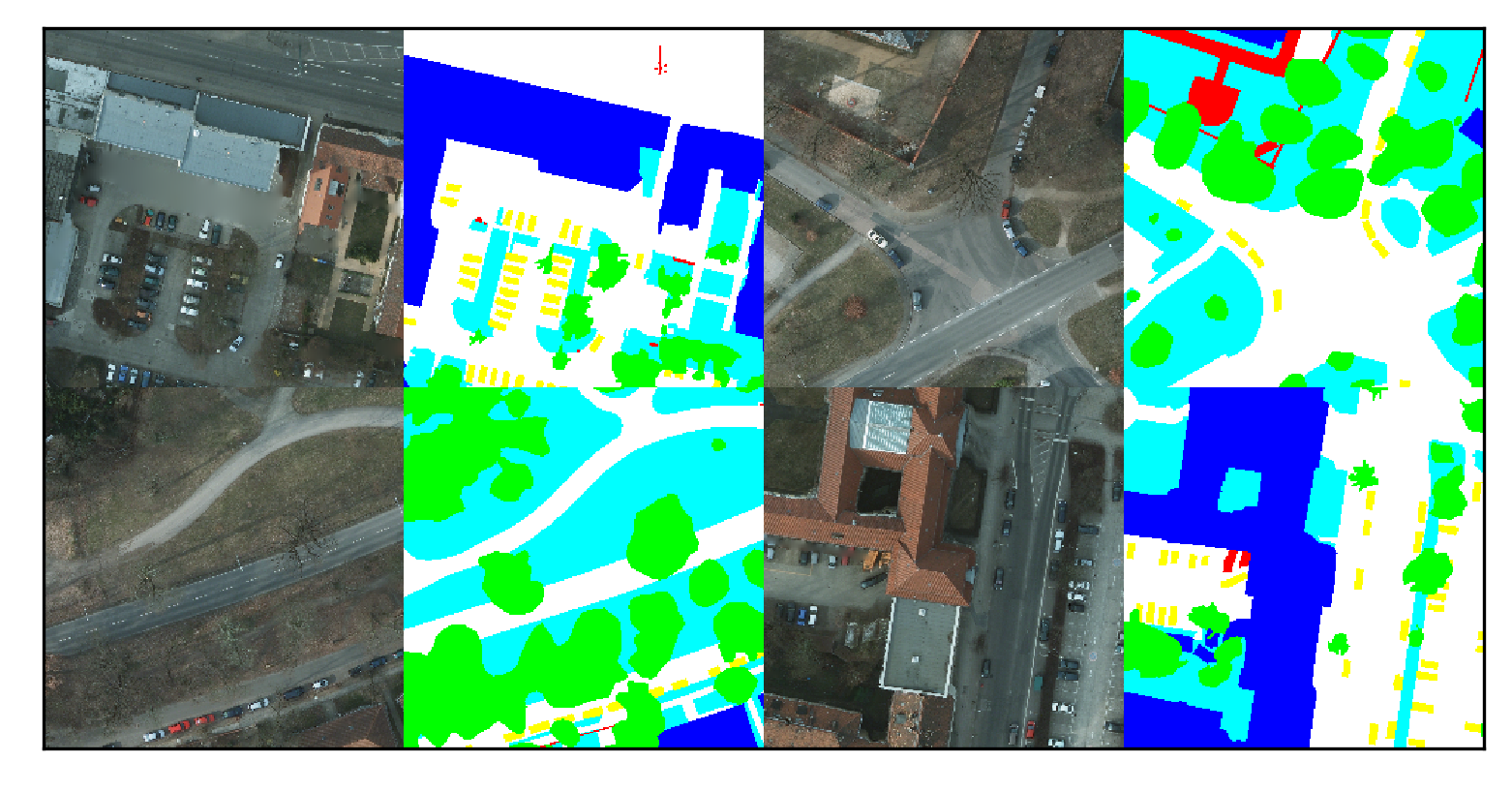}
\caption{Examples of 256 x 256 image chips with associated segmentation labels.}
\label{ISPRS_chip_example}
\end{figure}

For all modeling purposes, the images are chipped into 256 x 256 crops with a 32 pixel stride for semantic label modeling and object detection. For conditional image modeling purposes, random crops containing at least 10 vehicles are sampled from the full images in an attempt to reduce overfitting. Examples are presented in Figure \ref{ISPRS_chip_example}.

\subsection{Semantic Label Modeling}

To model and synthesize label distributions, we use the Progressive Growing of GANs (PGAN) work by Karras et al\cite{karras2017progressive}. The idea is to incrementally double the side length of the the output of the generator and the input of the discriminator during training by adding convolutional layers. At each stage, the side length is $2\textsuperscript{\textit{N}}$, where $\textit{N} \in \{2, 3, \dots, \log_2(M)\}$ and \textit{M} is the target height of a square input image. Incrementally increasing resolution benefits the training stability of both the discriminator and generator, where large scale features are initially modeled (global color variations) progressing to medium scale features (positions of objects and elements) and finally small scale features (local textures). To further increase training stability, Wasserstein distances and gradient penalties are used \cite{arjovsky2017wasserstein}.

The PGAN loss function is given by;

\begin{equation}
 \min_{G}  \max_{D} \sum \mathcal{L}_{\text{GAN}}(G,D) 
\end{equation}

Where $\mathcal{L}_{\text{GAN}}(G,D)$ is composed of;
\begin{subequations}
\begin{align}
  \text{\textbf{Discriminator}}& &\frac{1}{m}& \sum_{i=1}^{m} \bigg[    D(\mathbf{x}^{(i)}) - D(G(\mathbf{z}^{(i)})) \bigg]& \\
  \text{\textbf{Generator}}& &\frac{1}{m}& \sum_{i=1}^{m} \bigg[    D(G(\mathbf{z}^{(i)})) \bigg]
\end{align}
\end{subequations} where \textbf{z} is a latent vector sampled from the normal distribution given by $\mathcal{N}(0,\,\mathcal{I})$, \textbf{x} is real data, and $m$ is the batch size over which we average the loss.

The gradient penalty to aid model convergence is given by;

\begin{equation}
\lambda \mathbb{E}_{G(\mathbf{z})}[(\|\nabla_{G(\mathbf{z})}D(G(\mathbf{z}))\|_{2}-1)^2], 
\end{equation} where $\lambda$ is a weighting coefficient.

The PGAN methods have recently demonstrated a marked improvement in modeling high resolution images as measured by Fr\'{e}chet Inception Distance (FID) and Inception Score (IS) \cite{karras2017progressive, heljakka2018pioneer, arjovsky2017wasserstein, gulrajani2017improved}.
In addition, by incrementally growing the GAN the speed of network convergence is increased as more images, through greater batch sizes at low image resolutions, can be presented to the discriminator in less time.

\begin{figure}[htbp]
\centering
\includegraphics[width=150mm]{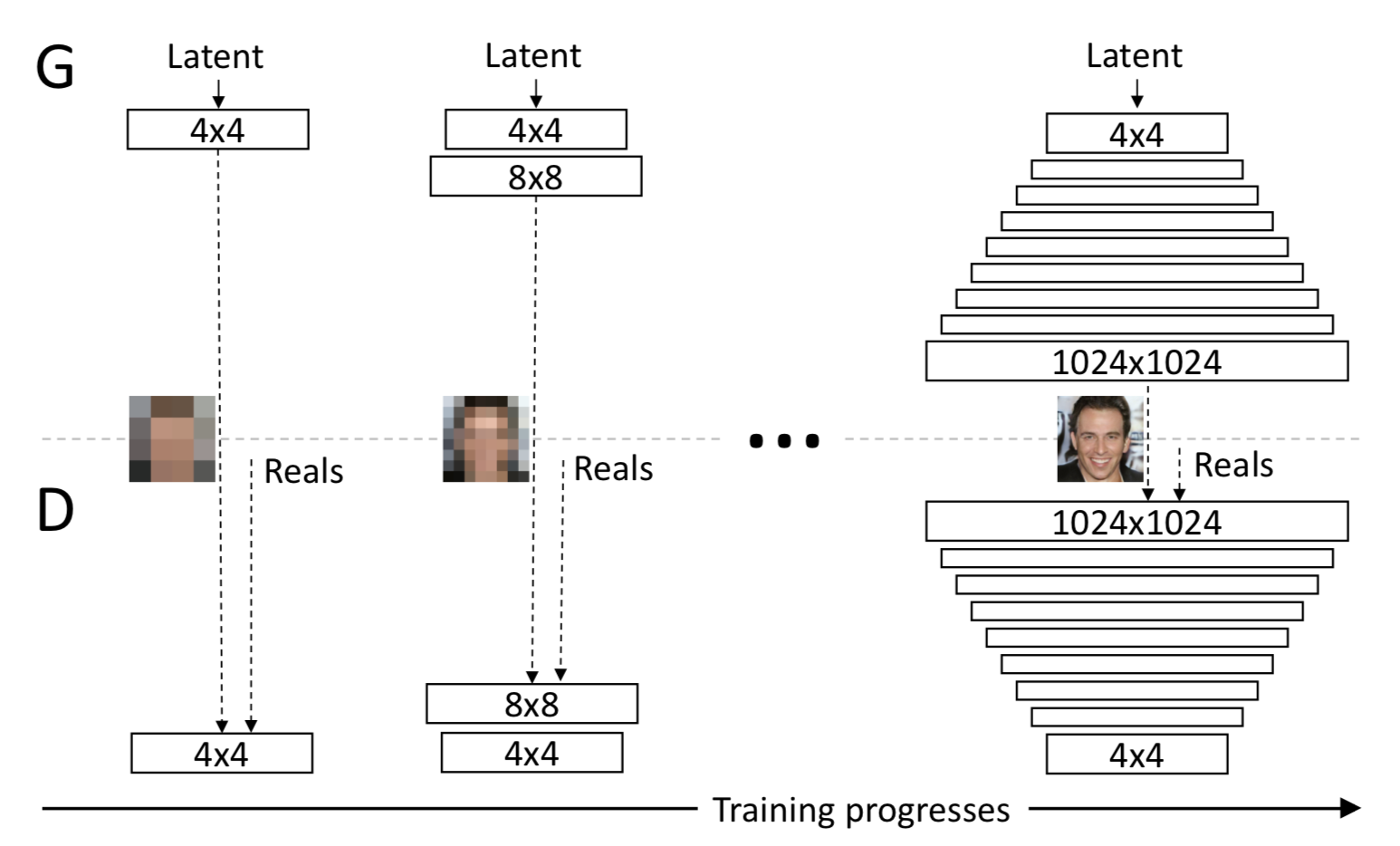}
\caption{PGAN architecture (from Karras et al \cite{karras2017progressive}).}
\label{pgan_architecture}
\end{figure}

We use the official TensorFlow implementation of PGANs Github repository \cite{karras2017progressiverepo, tensorflow2015-whitepaper}. Models are trained until 8M images are presented to the GAN. To increase network capacity, and thereby reduce the likelihood of mode collapse, we use a latent vector size of 1024. All other configurable variables are set to default values. In the repository's current implementation TensorFlow records are created from pre-cropped images. Therefore training time augmentation of the data set, other than mirroring and rotating, cannot be performed to help reduce overfitting. 

\subsection{Conditional Image Modeling}

To model images conditioned on segmentation labels, we use the semantically conditioned GAN (CGAN) work by Wang et al\cite{wang2018high}. This work improves upon the U-Net like architecture of the Pix2Pix method, upon which it is based, by introducing a multiscale coarse-to-fine generator and discriminator network which has a similarity to PGANs \cite{ronneberger2015u, isola2017image}. Here the objective of the generator is to transform segmentation labels to realistic synthetic images corresponding to these labels, while the discriminator attempts to distinguish real and synthetic images. 

The CGAN loss function is given by;

\begin{equation}
 \min_{G} \Bigg( \bigg( \max_{D_1,D_2,D_3} \sum_{k=1,2,3} \mathcal{L}_{\text{GAN}}(G,D_k) \bigg)  + \lambda \sum_{k=1,2,3} \mathcal{L}_{\text{FM}}(G,D_k) \Bigg) 
\end{equation} where $\mathcal{L}_{\text{GAN}}$ is the GAN portion of the loss function, which uses binary cross-entropy, not Wasserstein distance or gradient penalty as used in PGAN, and is given by:

\begin{subequations}
\begin{align}
  \text{\textbf{Discriminator}}& &\frac{1}{m}& \sum_{i=1}^{m} \bigg[    \log D_k(\mathbf{s}^{(i)},  \mathbf{x}^{(i)}) + \log(1 - D_k(\mathbf{s}^{(i)}, G(\mathbf{s}^{(i)}))) \bigg]& \\
  \text{\textbf{Generator}}& &\frac{1}{m}& \sum_{i=1}^{m} \bigg[    \log D_k(\mathbf{s}^{(i)},  G(\mathbf{s}^{(i)})) \bigg],
\end{align}
\end{subequations} and $\mathcal{L}_{\text{FM}}$ is the feature matching portion of the loss function, introduced to  map realistic textures and patches from real to synthetic imagery. Features are extracted from multiple layers of the discriminator when feeding forward real and synthetic images created by the generator. The loss between these feature representations is minimized. $\mathcal{L}_{\text{FM}}$ is given by;

\begin{equation}
 \mathbb{E}_{(\mathbf{s},\mathbf{x})} \sum_{i=1}^{T}\frac{1}{N_i}[\| D_{k}^{(i)}(\mathbf{s},\mathbf{x})-D_{k}^{(i)}(\mathbf{s},G(\mathbf{s}))\|_{1}]
\end{equation} where $\mathbf{s}$ is the semantic label, $k$ selects the discriminator scale, $T$ is the total number of layers, $N_i$ denotes the number of elements in each layer, and $\lambda$ is a weighting coefficient. 

CGAN also incorporates instance map information during training to differentiate objects of the same class that border and/or obscure each other. This allows the generation of synthetic images with plausible instances of the same class that obscure each other to have well-defined borders with different characteristics (color, texture, etc.). The Potsdam data set does not include such boundary information; however, it is uncommon for vehicles--our objects of interest to obscure one another due to the geometry of the image capture. Other objects, notably buildings, in the Potsdam data often border each other. We conjecture that our CGAN would produce higher quality synthetic images if given instance segmentations.

\begin{figure}[htbp]
\centering
\includegraphics[width=150mm]{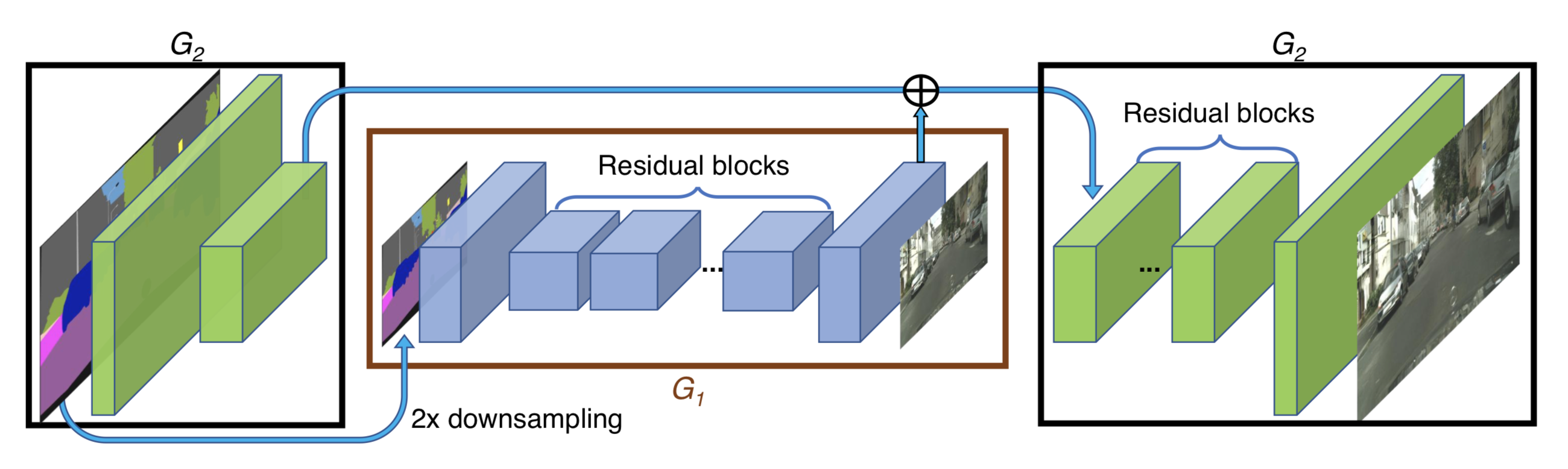}
\caption{CGAN architecture (from Wang et al \cite{wang2018high}).}
\label{pgan_architecture}
\end{figure}

The official PyTorch implementation Pix2PixHD Github repository is used \cite{wang2018highrepo, paszke2017automatic}. During training 256 x 256 crops are randomly chipped from Potsdam imagery with the associated label if 10 or more vehicles are present. This allows the generator and discriminator to be presented with randomized data in an attempt to reduce overfitting. In all cases models are trained to 50k steps. All other configurable variables are set to default values.

\subsection{Joint Modeling}

To generate synthetic image and label pairs first labels are generated using a trained PGAN model. Data is post-processed to select only synthetic labels which have distinct, well defined label values using a simple histogram test. After this stage label values are clustered and set to the nearest integer label value to match the six class labels in the Potsdam data. 

This post-processed synthetic label set is then fed through the trained CGAN model to create accompanying synthetic images. The images and label pairs are not post processed further. We study the efficacy of this joint modeling strategy to (1) produce plausible image / segmentation mask pairs as measured by FID (Section \ref{joint modeling}), and; moreover, (2) for use as a training data augmentation technique for a vehicle detection task (Section \ref{vehicle detection}). We decompose these studies as a function of available data.

\section{Experiments}
 \label{sec:experiments}

To determine the sensitivity of our joint modeling approach as a function of available data, we split the training data into groups containing few to many image chips. Each group is then used to train both PGAN and CGAN to produce synthetic image label pairs. An object detector, specifically a Feature Pyramid Network version of Single Shot Detector (FPN SSD) pretrained on the COCO data set, is then trained to detect the vehicle class using the TensorFlow object detection API \cite{liu2016ssd}. The amount of synthetic data added to the real data corpus is varied from zero to three times the number of real image chips in the training data set to determine the optimal benefits of synthetic data set augmentation.  The Common Objects in Context (COCO) metrics are used to measure detector accuracy, specifically mean average precision (mAP) @ 0.75 intersection over union (IoU) \cite{lin2014microsoft}. 
\subsection{Joint image and label modeling}\label{joint modeling} 

\begin{figure}[htbp]
\centering
\includegraphics{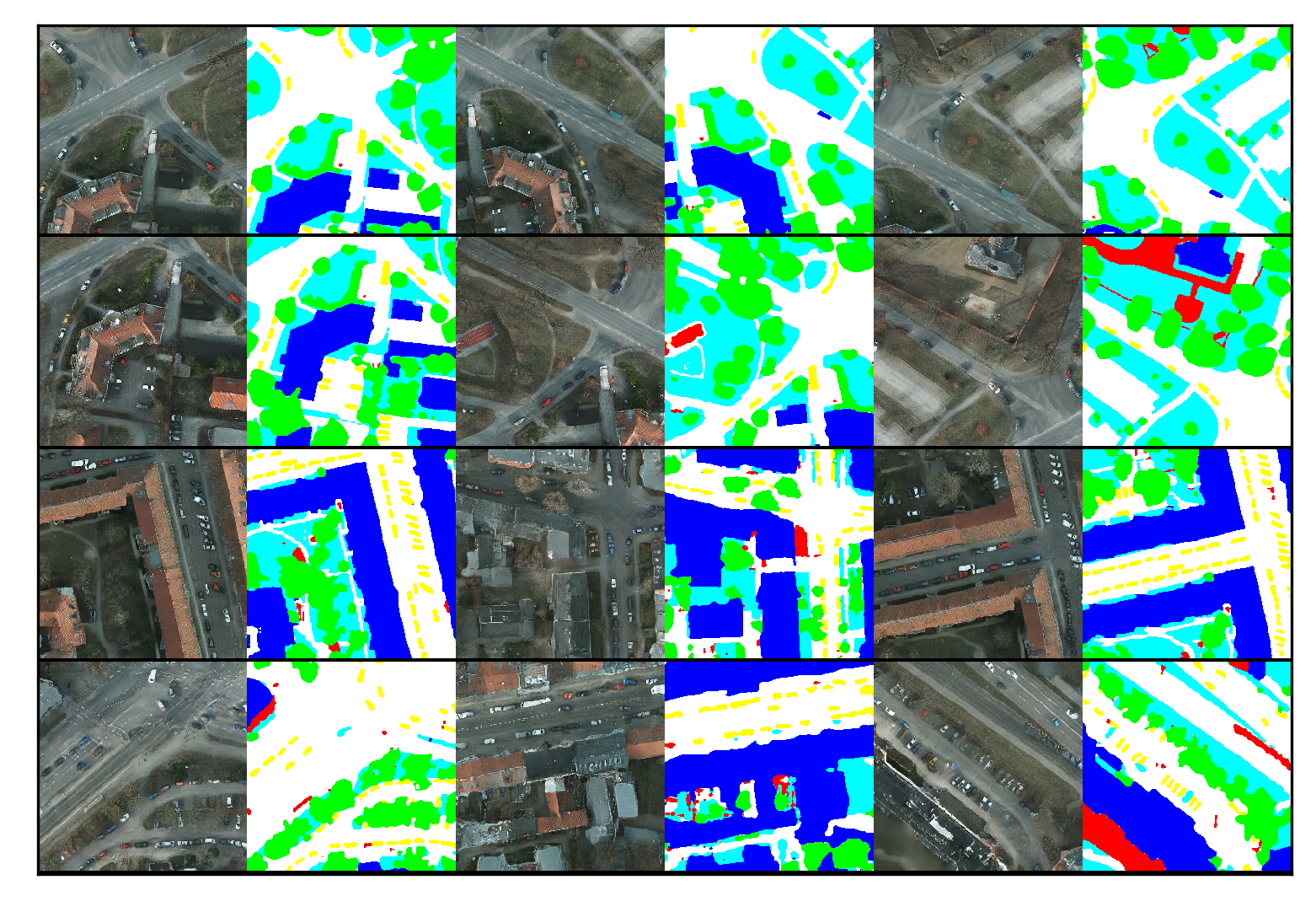}
\caption{Examples of synthesized image and label pairs for varying amounts of image chips presented. The number of chips and vehicles presented in the four rows from top to bottom are (chip number/vehicle number); row 1  60/56, row 2 203/103, row 3 249/210, row 3 919/513, and row 4 1591/1130.}
\label{joint_array}
\end{figure}

Figure \ref{joint_array} presents examples of synthesized image and label pairs for varying amounts of image chips presented. The number of chips and vehicles presented in the four rows from top to bottom are (chip number/vehicle number); row 1  60/56, row 2 203/103, row 3 249/210, row 3 919/513, and row 4 1591/1130. 

It is evident that the diversity of the image label pairs increases as a function of number of chips presented to the GAN. This is to be expected given that the GAN is modeling the underlying distribution and variation of the data set presented to it. The model can only interpolate within the data set distribution, not extrapolate entirely different scenes. In addition, when few chips are used for modeling overfitting is present. Here the same, or very similar, patterns are reproduced which closely mimic the real data set.   

When the number of training chips is increased the synthetic data diversity increases. When large amounts of training chips are available the synthetic labels, although still subjectively realistic, exhibit distortion, where boundaries are no longer straight or regular. The amount of variation in the ISPRS Potsdam data set is much larger than that of CelebA and LSUN as used in the original PGAN paper \cite{yu15lsun, liu2015deep}. In order to create highly realistic celebrity faces a sequence of data preprocessing and filtering is needed before modeling, such as registration of key points and having a common image resolution.  

\begin{figure}[htbp]
\centering
\includegraphics[width=150mm]{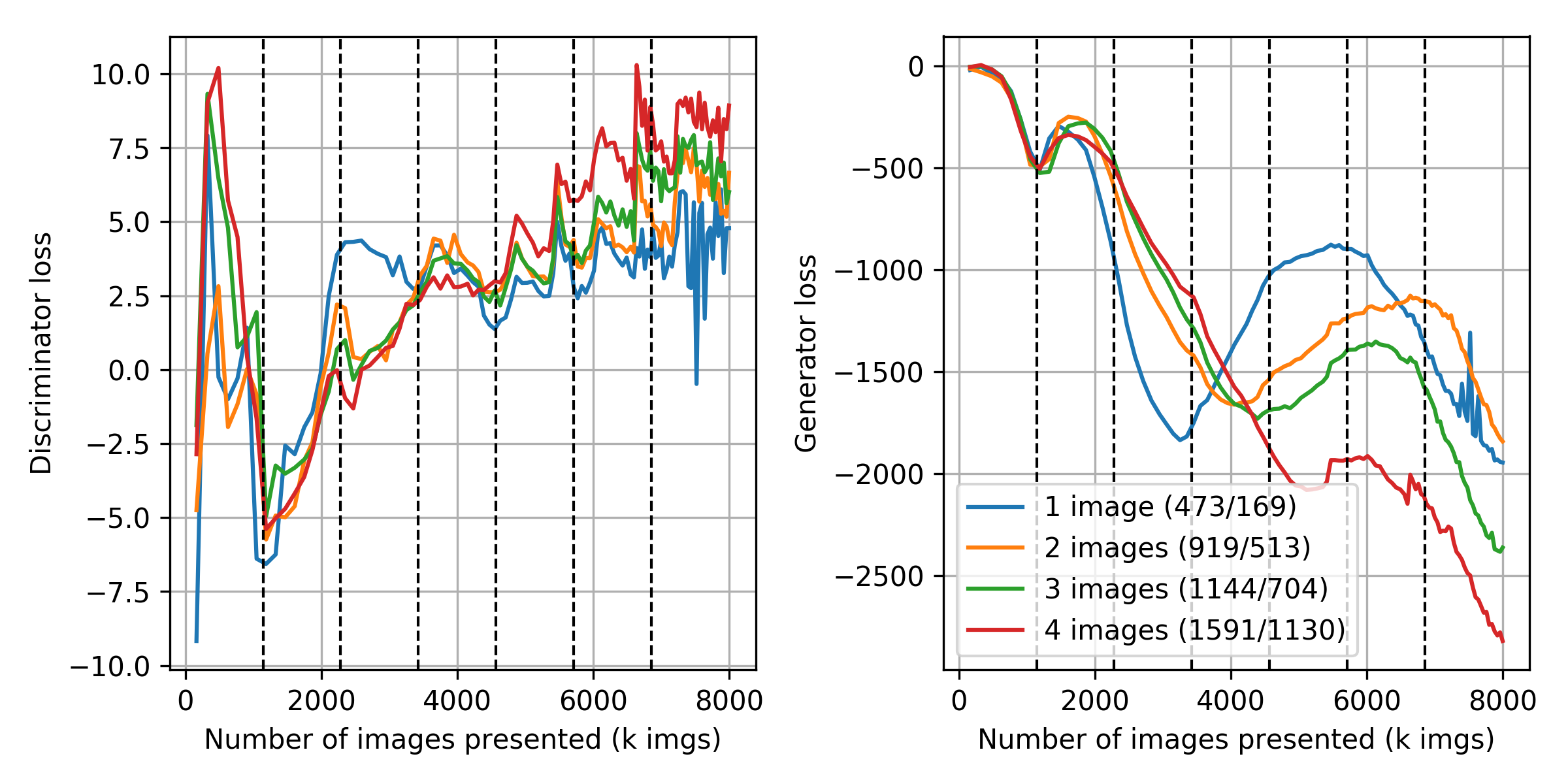}
\caption{PGAN discriminator and generator loss as a function of number of images used in training. Vertical dashed lines denote where image resolution doubles.}
\label{loss_graph}
\end{figure}

Figure \ref{loss_graph} presents the discriminator (\textit{D}) and generator (\textit{G}) loss of PGAN during model training. Losses closer to zero represent a more accurate model; \textit{D} is able to discriminate real and synthetic data, while synthetic data created by \textit{G} is able to fool \textit{D}. As the resolution of the data being presented is increased during training, the effectiveness of the generator to fool the discriminator decreases. This is typical when training PGAN; creating high resolution synthetic data is a much more difficult task than creating low resolution data.  

When the number of images, and the variation in the images, available to train PGAN increases the absolute value of the generator loss at high data resolutions increases. This suggests that the output of the generator decreases in image quality such that the discriminator can differentiate real and synthetic data with more accuracy. Another method to determine the image quality of the synthetic data is to calculate the Fr\'{e}chet Inception Distance (FID) \cite{heusel2017gans}. Here the 2048 dimensional activations of the Inception-v3 pool3 layer, trained on the ImageNet data set, are extracted between real and synthetically generated data \cite{Dengimagenet}. These two distributions are then compared using the Fr\'{e}chet distance, where lower FID suggests more realistic synthetic data, as shown in the following equation;

\begin{equation}
  \text{FID} = \| \mu_\text{real} - \mu_\text{synthetic} \|_2 + \text{Tr}(\Sigma_\text{real} + \Sigma_\text{synthetic} - 2 (\Sigma_\text{real} \Sigma_\text{synthetic})^{1/2} )  \
\end{equation}

Calculated FID between real and synthetic data as a function of number of images used in training is shown in  table \ref{fid}. There is a trend towards greater FID when the number of images available for training is increased. This suggests that larger data sets cannot be modeled as effectively when compared to smaller data sets which exhibit less variation.

\begin{table}[htbp]
\caption{FID as a function of number of images used in training.} 
\label{fid}
\begin{center}       
\begin{tabular}{|c|l|} 
\hline
\rule[-1ex]{0pt}{3.5ex}   \textbf{Number of images / vehicles} & \textbf{FID}  \\
\hline
\rule[-1ex]{0pt}{3.5ex}  1 / 169 & 60.14  \\
\hline
\rule[-1ex]{0pt}{3.5ex}  2 / 513 & 65.44   \\
\hline
\rule[-1ex]{0pt}{3.5ex}  3 / 704 & 90.88   \\
\hline
\rule[-1ex]{0pt}{3.5ex}  4 / 1130 & 118.72   \\
\hline
\end{tabular}
\end{center}
\end{table}

Similar effects are also observed in the accompanying synthetic imagery. For small amounts of chips object texture is subjectively extremely realistic, most likely due to overfitting. As the number of training chips increases the textures begin to merge for different object classes. For example some vehicles and building rooftops have multiple colors and textures. 

\subsection{Vehicle Detection}\label{vehicle detection}

To evaluate the utility of our pipeline, we trained a standard object detection network\textemdash the Feature Pyramid Network version of Single Shot Detector (FPN SSD) pretrained on the COCO data set using the TensorFlow object detection API. The network was trained on all combinations of four nested sets of data (corresponding to the rows of Figure \ref{joint_array}) and seven ratios of real to synthetic data ranging from a baseline of 0\% synthetic data to the extreme case of 300\% synthetic data or three times as much synthetic data as real data. Each of the 28 models was trained for 40k epochs using the default hyper-parameters provided by the TensorFlow object detection API, including the standard $D_4$ augmentations of 90 degree rotations and mirroring. All training sessions appeared to have converged and showed no obvious signs of over-fitting.

\begin{figure}[htbp]
\centering
\includegraphics[width=150mm]{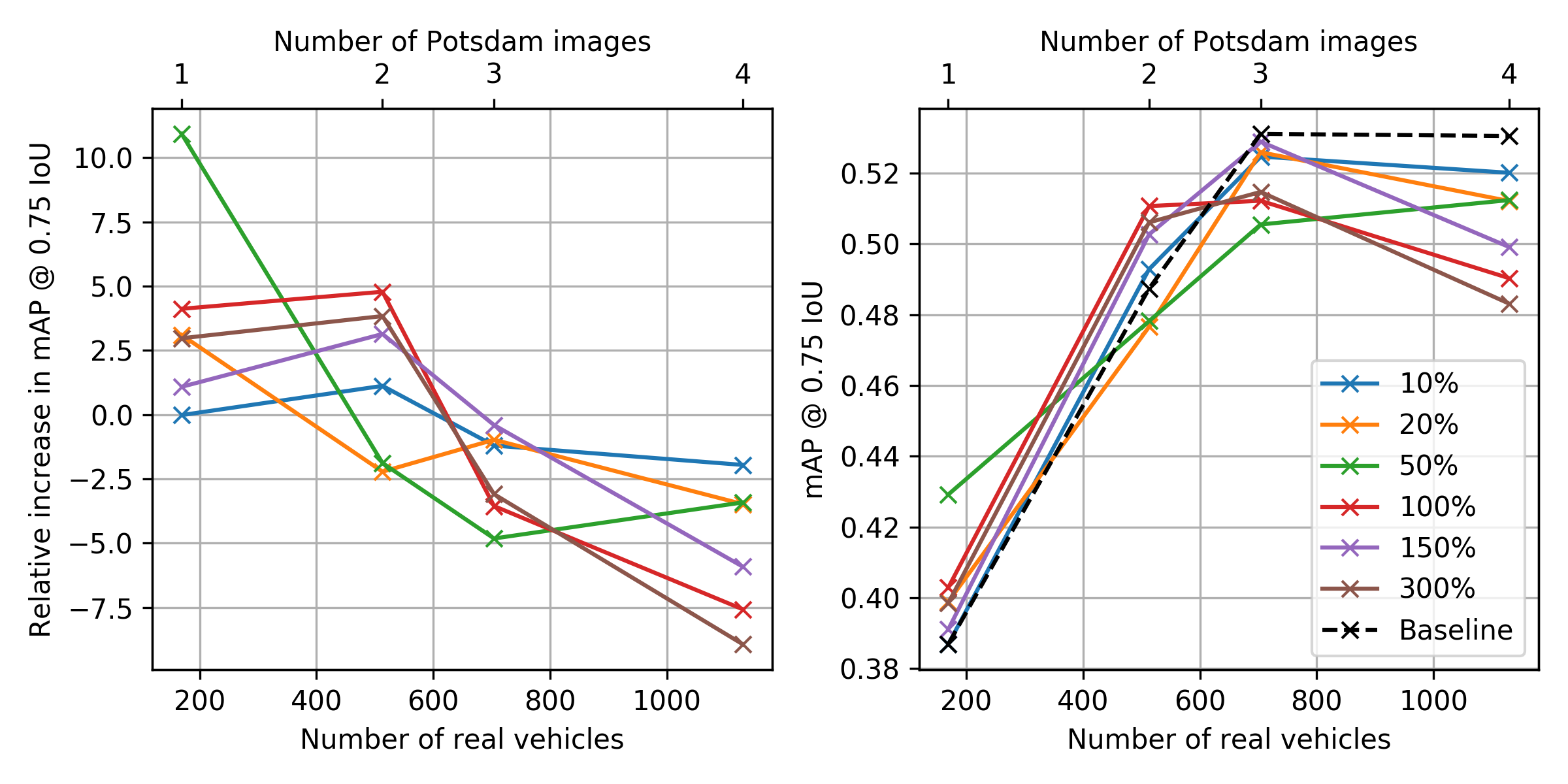}
\caption{Absolute and relative change in mAP @ 0.75 IoU as a function of number of real vehicles in the training data.  Synthetic data is added to the real data corpus by a relative amount.}
\label{rel_plot}
\end{figure}

We evaluated the detection networks using standard COCO metrics. Figure \ref{rel_plot} shows the absolute and relative change in mAP and average recall as a function of number of real vehicles in the training data. Synthetic data is added to the real data corpus by a relative amount.

Ideally, the relative benefit of the synthetic data would be positive but converge downwards to zero as the corpus of real data grows. While we find a consistent, positive benefit for small data sets and a downward trend as the data sets grow, the synthetic samples actually hurt performance for the larger two data sets. We hypothesize that this trend is connected with the observation from Table \ref{fid} that the FID scores correlate with training set size\textemdash with a richer corpus of real data, it becomes easier for the generator to “fool” the discriminator with data that is actually out-of-domain and thus ultimately unhelpful for our target task of detection. The relative increase in mAP for small dataset sizes is consistent with literature \cite{lesort2018evaluation}. 

To overcome this problem, one might either increase the capacity of the networks in our pipeline, condition the GAN on the image identity, add an autoencoder to the GAN, or build a synthesis pipeline that tries to optimize for improved detector performance directly. We hope to consider the latter approach in future work. 

\section{Summary}

A synthetic data pipeline is built to augment an object detection training corpus of remote sensed imagery. The results demonstrate that in low-sample scenarios a data-driven, GAN-based pipeline can enlarge the effective size of a corpus to improve detector performance beyond standard data augmentation techniques. The technique provides a consistent increase in mAP for smaller datasets, exhibiting a greater than 10\% relative increase when using a single Potsdam image. This observed maximum increase occurs when adding an additional 50\% of synthetic data to the real data corpus. The benefit of adding synthetic data reduces for greater amounts of additional synthetic data. When using two Potsdam images the increase in mAP is generally observed when using an additional 100\% of synthetic data or more. Here the maximum observed increase in mAP is 5\%, which is still a reasonable amount. In other words, given one or two Potsdam training images, the technique improves detector performance about half as much as adding an additional training image would. Practitioners will need to consider their problem's individual circumstances such as data availability, labeling costs, and sensitivity to performance gains when deciding whether to use the technique. 

While the technique provides a consistent increase in mAP for smaller training sets, the benefit decreases as the corpus grew and actually hurt performance on larger data sets. We believe that this problem is due to the generative pipeline's inability to model effectively the high variability found in the larger corpus. This is reflected in the generator training loss and the FID values when increasing the amount of training data. We believe that conditioning PGAN on the image identity may help counteract this. In future work, we hope to explore an end-to-end approach that unifies the segmentation and image synthesis tasks and directly optimizes for improved detection performance rather than just trying to sample from the distribution of real data.

\acknowledgments     
 
 This work was supported by NGA / NVIDIA Cooperative Research and Development Agreement \newline HM0476CRFY17007 and NRO / CACI contract 12-D-0227. It is approved for public release by National Geospatial Intelligence Agency \#19-862.


\bibliography{PCGAN_SPIE}   
\bibliographystyle{spiebib}   

\newpage
\appendix
\section{PGAN examples}\label{sec:appendixpganexamples}
\begin{figure}[htbp]
\centering
\includegraphics[width=150mm]{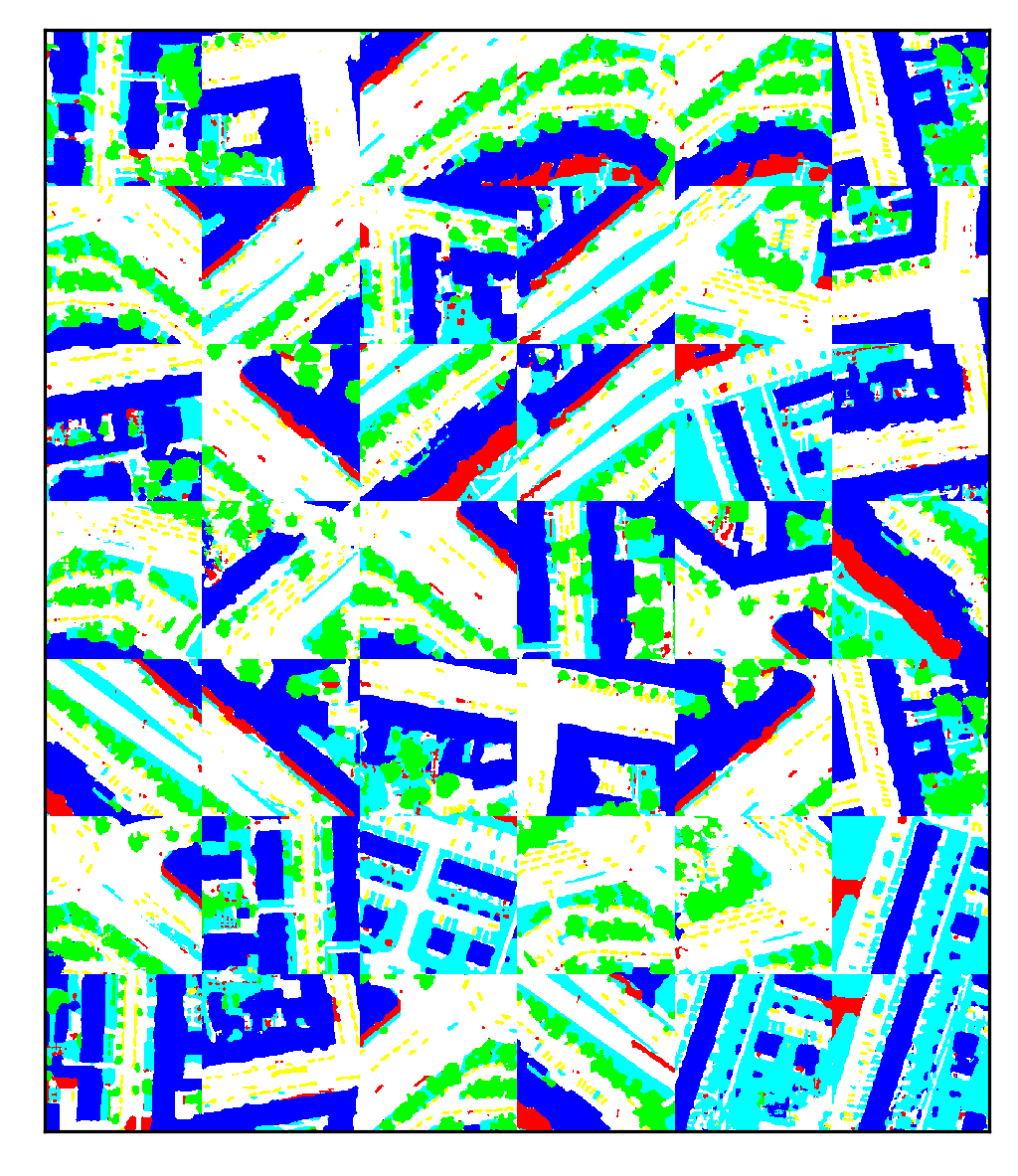}
\caption{Examples of synthetic labels generated by PGAN after filtering using a simple color histogram test.}
\label{pgan_examples_appendix}
\end{figure}

\newpage
\section{Joint PGAN and CGAN examples}\label{sec:appendixcganexamples}
\begin{figure}[htbp]
\centering
\includegraphics[width=150mm]{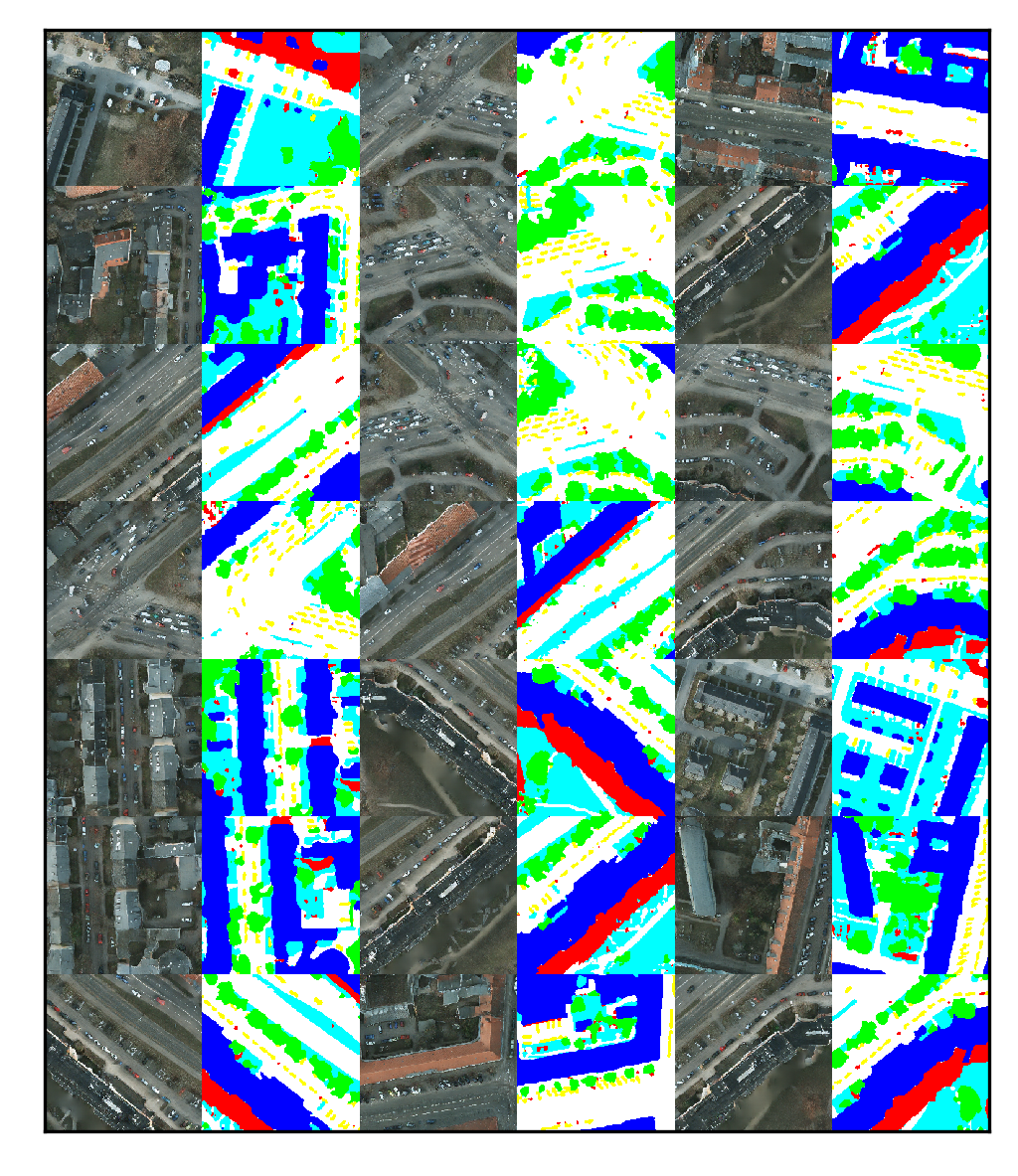}
\caption{Examples of synthetic images and accompanying labels generated by PGAN and CGAN after filtering using a simple color histogram test.}
\label{cgan_examples_appendix}
\end{figure}

\end{document}